\def\year{2016}\relax
\documentclass[letterpaper]{article}
\usepackage{aaai16}
\usepackage{times}
\usepackage{helvet}
\usepackage{courier}

\usepackage{algorithm}
\usepackage{algpseudocode}
\usepackage{graphicx}
\usepackage{amssymb}
\usepackage[font=scriptsize]{subcaption}
\usepackage[font=scriptsize]{subfig}
\usepackage{multirow}
\usepackage{float}
\newcommand*{\QEDB}{\hfill\ensuremath{\square}}

\begin{document}

\title{Path Planning in Dynamic Environments with Adaptive Dimensionality}

\author{Anirudh Vemula \and Katharina Muelling \and Jean Oh\\
Robotics Institute \\
Carnegie Mellon University\\
avemula1@andrew.cmu.edu, \{kmuelling, jeanoh\}@nrec.ri.cmu.edu}

\maketitle
\begin{abstract}
Path planning in the presence of dynamic obstacles is a challenging problem due to the added time dimension in the search space. In approaches that ignore the time dimension and treat dynamic obstacles as static, frequent re-planning is unavoidable as the obstacles move, and their solutions are generally sub-optimal and can be incomplete. To achieve both optimality and completeness, it is necessary to consider the time dimension during planning. The notion of adaptive dimensionality has been successfully used in high-dimensional motion planning such as manipulation of robot arms, but has not been used in the context of path planning in dynamic environments. In this paper, we apply the idea of adaptive dimensionality to speed up path planning in dynamic environments for a robot with no assumptions on its dynamic model. Specifically, our approach considers the time dimension only in those regions of the environment where a potential collision may occur, and plans in a low-dimensional state-space elsewhere. We show that our approach is complete and is guaranteed to find a solution, if one exists, within a cost sub-optimality bound. We experimentally validate our method on the problem of 
3D vehicle navigation (x, y, heading) in dynamic environments. 
Our results show that the presented approach achieves substantial speedups in planning time over 4D heuristic-based A*, especially when the resulting plan deviates significantly from the one suggested by the heuristic.
\end{abstract}


\section{Introduction}
\label{intro}

It is important for mobile robots to be able to generate collision-free paths in environments that contain both static and dynamic obstacles.
In static environments, robots can efficiently generate a collision-free path using the occupancy gridmap of the environment. But in dynamic environments, to account for the dynamic nature of obstacles, the robot needs to predict the future trajectories of these obstacles to plan its own path accordingly. These predictions involve a high degree of uncertainty due to sensor limitations and incorrect dynamic models. As a result, the predicted trajectories are subject to frequent changes due to incorrect predictions, which makes it necessary to generate new plans in a timely manner.

To account for dynamic obstacles in an environment, we need to include the time dimension into consideration. 
For example, planning a path for a non-holonomic robot in a dynamic environment involves a 4D state-space, i.e., ($x, y,$ heading, time). 
Due to the curse of dimensionality, adding the time dimension substantially increases the number of states to be searched, e.g., from 3D state-space considering only ($x, y,$ heading), leading to long planning times especially, since there are potentially an unbounded number of timesteps for each spatial location.

The Adaptive Dimensionality (AD) approach, \cite{gochev2011path}, exploits the observation that while planning in a high dimensional space is needed to satisfy kinematic constraints and collision-free criteria, large portions of the path are still low dimensional. For instance, in planning for a non-holonomic robot, an optimal path generally includes straight-line segments that do not involve any turns or collisions with dynamic obstacles. This observation implies that high dimensional path planning is required only in the sections of the path where turning is required or where there is a potential collision with a dynamic obstacle. 

Following this insight, we consider the time dimension only in those regions where a potential collision could occur and ignore it elsewhere. In this paper, we develop an approach that can achieve speedups over full-dimensional heuristic-based A* without any assumptions on robot capabilities by employing a variant of the Adaptive Dimensionality approach.

In the remainder of this paper, we will give an overview of relevant existing work in Section \ref{related-work}. 
The planning problem is formally defined in Section \ref{sec:problem} and the motivation for our approach is presented in Section \ref{sec:motivation}.
Sections \ref{sec:approach} and \ref{theory} will describe our approach and prove the theoretical guarantees. The efficiency of the method is demonstrated by applying it to a 3D non-holonomic robot navigation problem in the presence of dynamic obstacles, showing a significant increase in speed over 4D heuristic-based A* planner for this task, in Sections \ref{exp} and \ref{results}.


\section{Related Work}
\label{related-work}
Our work is relevant to path planning in dynamic environments and works on coping with high dimensionality. In general, we divide the existing approaches into three categories: works that deal with planning in dynamic environments, that deal with high-dimensional planning using adaptive dimensionality and that use hybrid dimensionality path planning in dynamic environments.

\subsection{Path Planning in Dynamic Environments}

A common approach used for efficient path planning in dynamic environments involves modeling moving obstacles as static objects with a small window of high cost around the beginning of their projected trajectories~\cite{likhachev2009planning,rufli2009smooth}. 

By avoiding the additional time dimension,
these approaches can efficiently find paths that do not collide with any obstacles in the near future. However, they can suffer from severe sub-optimality or even incompleteness due to the uncertainty of moving obstacles in the future.

To plan and re-plan online, several approaches have been suggested that sacrifice near-optimality guarantees for efficiency~\cite{van2006anytime}, including sampling-based planners such as RRT-variants that can quickly obtain kinodynamically feasible paths in a high dimensional space~\cite{bekris2007greedy,petti2005safe}. However, these sampling-based approaches do not provide any global optimality guarantees that we require in most cases. 

Other approaches \cite{fox1997dynamic,brock1999high} delegate the dynamic obstacle avoidance problem to a local reactive planner which can effectively avoid collision with dynamic obstacles. These methods have the disadvantage that they can get stuck in local minima and are generally not globally optimal. 

Among works that provide global optimality guarantees that are relevant to our work, HCA* ~\cite{silver2005cooperative} is an approach that plans in the full space-time search space for a path from start to goal, taking dynamic obstacles into account under the guidance of a low-dimensional heuristic. In dynamic environments, HCA* provides guarantees on optimality and can be applied to path planning for a robot without any assumptions on its motion model.

Recently, approaches such as SIPP and its variants~\cite{phillips2011sipp,narayanan2012anytime}, have been introduced that obtain fast, globally optimal paths in dynamic environments. But, SIPP assumes that the robot is capable of waiting in place. 
In cases where this assumption doesn't hold, SIPP is essentially a full space-time A* planner. 
Thus, the advantages from this algorithm are restricted to only those robots which have the capability of waiting in place, unlike a fixed wing aircraft or a motorcycle. 
Besides, when fuel efficiency is included in the cost, fuel consumption is generally higher during idling than moving. 

We use HCA* as our baseline algorithm as it doesn't make any assumptions on the motion model of the robot and provides optimality guarantees, similar to our approach. 

\subsection{Adaptive Dimensionality}
To accelerate planning, a variety of algorithms try to avoid global planning in high-dimensional state-space. 
In these algorithms, planning is split into a two-layer process where a global planner deals with a low-dimensional state-space and provides an input to a high-dimensional local planner~\cite{philippsen2003smooth}. The local planner is a reactive planner that avoids obstacles locally and hence, is fast and efficient. However, these approaches can result in paths that are highly suboptimal or that cannot be executed, due to mismatches in the assumptions made by the global and local planners. 

In~\cite{knepper2006high}, highly accurate heuristic values are computed by solving a low-dimensional problem and are then used to direct high-dimensional planning. However, this approach does not explicitly decrease the dimensionality of the state-space and can lead to long planning times when the heuristic is incorrect. 

By contrast, the Adaptive Dimensionality (AD) approach, \cite{gochev2011path}, explicitly decreases the dimensionality of the state-space in regions where full-dimensional planning is not needed. This approach introduces a strategy for adapting the dimensionality of the search space to guarantee a solution that is still feasible with respect to a high dimensional motion model while making fast progress in regions that exhibit only low-dimensional structure. In \cite{gochev2012planning}, path planning with adaptive dimensionality has been shown to be efficient for high-dimensional planning such as mobile manipulation. The AD approach has been extended in~\cite{gochev2013incremental}, to get faster planning times by introducing an incremental planning algorithm. \cite{zhang2012combining} extends this method in the context of mobile robots by using adaptively dimensional state-space to combine the global and local path planning problem for navigation. Our approach builds on the AD approach and applies it to path planning in dynamic environments.

\subsection{Hybrid Dimensionality in Dynamic Environments}
Some approaches only plan in full-dimensional space-time search space until the end of an obstacle's trajectory and then finish the plan in a low-dimensional state-space. Time-bounded lattice planning, \cite{kushleyev2009time}, neglects dynamic obstacles and the time dimension in the search space after a certain point in the time. Several works, \cite{petereit2013mobile,petereit2014combined}, have extended this algorithm to account for kinematic and dynamic feasibility in the resulting paths by using a hybrid dimensionality state-space. These approaches sacrifice optimality for faster planning times and don't provide theoretical guarantees on the sub-optimality of the solution. In addition, our algorithm doesn't prune the dynamic obstacle trajectories, instead takes the entire obstacle trajectories into account and returns a bounded sub-optimal collision-free path. Considering the entire trajectory of the obstacles ensures a globally optimal solution.


\section{Problem Definition}\label{sec:problem}

In this paper, we follow the simplifying assumptions used in~\cite{phillips2011sipp} that the trajectories of moving obstacles are known and that obstacles move at a constant speed. Based on these assumptions, path planning in a dynamic environment can be formulated more generally as path planning in a high-dimensional space as follows. A path planning problem is defined as a tuple $\Phi = [G = (S, T), c, X_{s}, X_{g}]$, where $G$ denotes a graph consisting of $S$, a set of discretized states in a $d$-dimensional space, and $T$, a set of feasible transitions between each pair of states $X_i, X_j \in S$; $c$, a function encoding a non-negative cost $c(X_i,X_j)$ for each pair of transitions $(X_i,X_j) \in T$; $X_{s} \in S$, a start state, and $X_{g} \in S$, a goal state. For instance, the target problem for a ground vehicle can be defined in 4-D state space $(x, y, \theta, t)$ where each variable denotes $x$-coordinate, $y$-coordinate, vehicle heading, and time, respectively. Note that transitions that result in collision with an obstacle are assigned infinite cost, making them invalid. 

A path between states $X_i$ and $X_j$ is denoted by $\pi(X_i,X_j)$, and the cost of a path is defined as the sum of all transition costs in the path. 

Given a planning problem $\Phi = [G, c, X_{s}, X_{g}]$, the goal is to find a minimum cost path between the two states $X_{s}$ and $X_{g}$, denoted by $\pi^*(X_s,X_g)$. 
Alternatively, given a suboptimality bound $\epsilon$, the goal of the planner can be relaxed to find a path $\pi(X_S,X_G)$ such that its cost $c(\pi(X_S,X_G)) \leq \epsilon \cdot c(\pi^*(X_S,X_G))$.


\section{Motivation}\label{sec:motivation}

Given a path planning problem in a high dimensional space, it is possible to find an optimal solution through a complete search. For example, heuristic-based A* variant algorithms exist that are guaranteed to find an optimal solution~\cite{silver2005cooperative}. Because these algorithms rely on low-dimensional heuristics, search can be counter-intuitive.  

\begin{figure}[tb!]
\centering%
\includegraphics[scale=0.4]{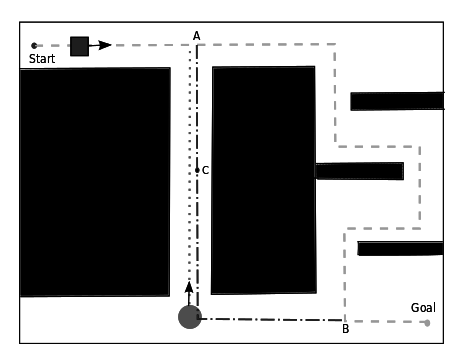}%
\caption{Example of a dynamic environment where the heuristic leads the robot (square) into collision (on dash-dot path) with the dynamic obstacle (disc) at C. We need to find an alternate path (dashed path) from A to B without expanding a large number of states.}%
\label{fig:intro}
\end{figure}

Consider the example shown in Figure \ref{fig:intro}, where the resulting path (dash-dot path), in the absence of the dynamic obstacle (disc), is towards the heuristic. But, in the presence of the dynamic obstacle, this path is in collision and cannot be executed. Hence, we need to come up with the alternative path (dashed path), which is against the heuristic. Heuristic-based A* would expand a large number of states and will take a long time to generate the new path whereas our approach generates the alternative path quickly, because it plans in a lower dimensional space.

More generally, we observe that substantial sections of paths found are not in collision with any dynamic obstacles, implying that we need not consider the time dimension in such regions. We can obtain quicker planning times by planning in low dimensional state-space for those regions and in full dimensional state-space only where it is necessary to reason about a potential collision with an obstacle. 

Based on these observations, we explore the idea of adaptive dimensionality to solve the target problem.


\section{Approach}\label{sec:approach}

We describe the adaptive dimensionality approach used for path planning in dynamic environments, and the algorithm for finding a bounded cost sub-optimal path.

\subsection{Adaptive Dimensionality for Dynamic Environments}
Our approach follows the algorithm for planning with adaptive dimensionality introduced in~\cite{gochev2011path}.
Following their notation, the target problem in Section~\ref{sec:problem} can be rewritten as follows.
Graph $G$ is substituted with the adaptive-dimensionality graph $G^{ad} = (S^{ad}, T^{ad})$. $G^{ad}$ is constructed from two graphs: a high dimensional graph $G^{hd} = (S^{hd}, T^{hd})$ with dimensionality $h$ and a low dimensional graph $G^{ld} = (S^{ld}, T^{ld})$ with dimensionality $l$. The state-space $S^{ld}$ is a projection of $S^{hd}$ onto a lower dimensional manifold ($ h \textgreater l$) through a projection function:
\begin{equation}
\lambda : S^{hd} \rightarrow S^{ld}
\end{equation}\label{eq:lambda}
Similarly, an inverse projection function $\lambda^{-1} : S^{ld} \rightarrow \mathcal{P}(S^{hd})$ is defined to map low-dimensional states to the set of all their high-dimensional pre-images,where $\mathcal{P}(S^{hd})$ denotes the power set of $S^{hd}$.

Both state spaces $S^{hd}$ and $S^{ld}$ can have their own transition sets $T^{hd}$ and $T^{ld}$,
with a constraint that transitions in a high-dimensional space are more expensive than the corresponding transitions in a low-dimensional space, that is for every pair of states $X_i$ and $X_j$ in $S^{hd}$,
\begin{equation}
c(\pi^*_{G^{hd}}(X_i,X_j)) \geq c(\pi^*_{G^{ld}}(\lambda(X_i), \lambda(X_j)))
\label{eq:cost}
\end{equation}
We note that this constraint is important for bounding the suboptimality that will be discussed later in Section~\ref{theory}.

\begin{figure*}[ht]
  \centering
  \begin{subfigure}[t]{0.3\linewidth}
    \centering
    \includegraphics[width=\linewidth]{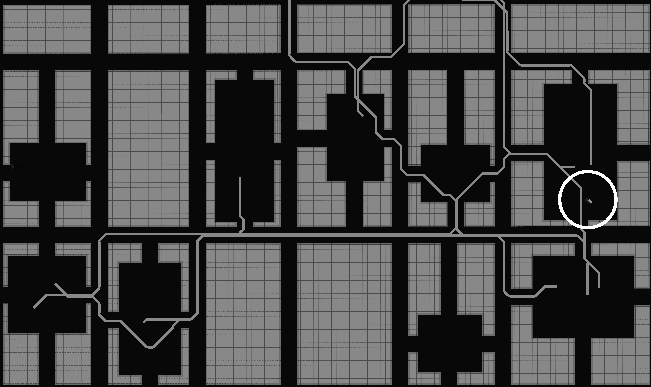}
    \caption{HD region at the start}
  \end{subfigure}
  ~
  \begin{subfigure}[t]{0.3\linewidth}
    \centering
    \includegraphics[width=\linewidth]{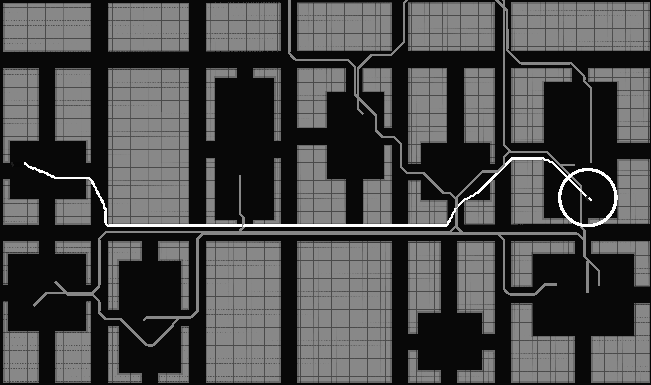}
    \caption{Path returned by planning phase in first iteration}
  \end{subfigure}
  ~
  \begin{subfigure}[t]{0.3\linewidth}
    \centering
    \includegraphics[width=\linewidth]{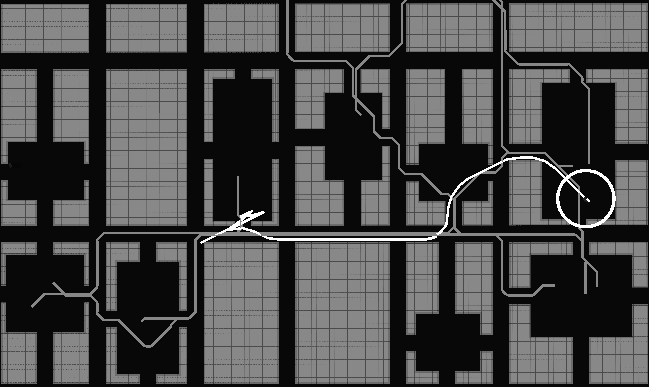}
    \caption{Search cannot progress in tunnel due to collision}
  \end{subfigure}
  
  \begin{subfigure}[t]{0.3\linewidth}
    \centering
    \includegraphics[width=\linewidth]{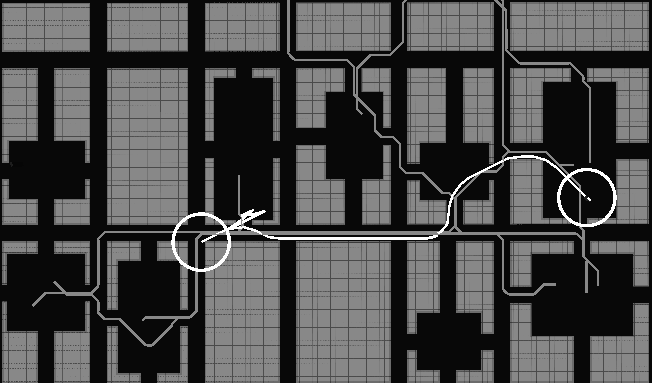}
    \caption{HD region introduced at point of collision}
  \end{subfigure}
  ~
  \begin{subfigure}[t]{0.3\linewidth}
    \centering
    \includegraphics[width=\linewidth]{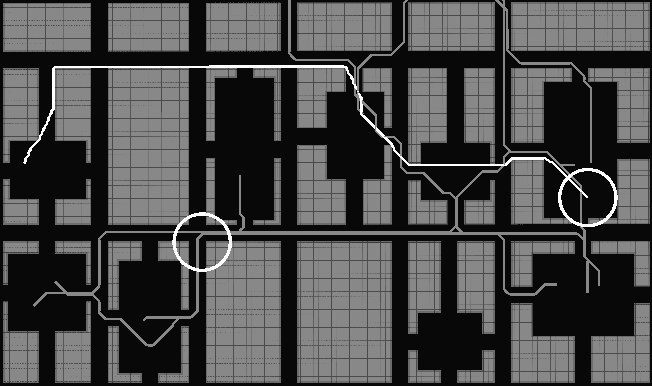}
    \caption{Path returned by planning phase in second iteration}
  \end{subfigure}
  ~
  \begin{subfigure}[t]{0.3\linewidth}
    \centering
    \includegraphics[width=\linewidth]{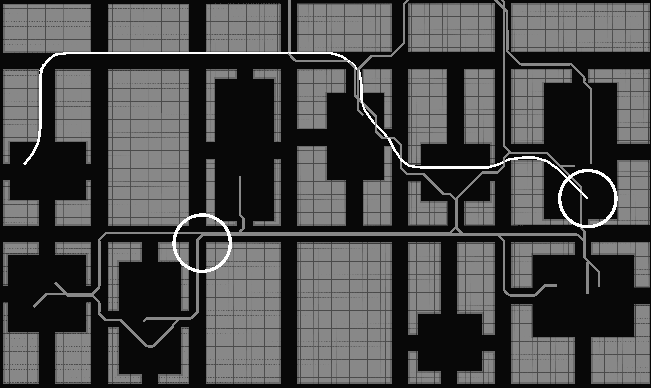}
    \caption{Tracking successful and path returned as solution}
  \end{subfigure}

  \caption{Example run of the algorithm on a sample map. HD regions are indicated by white circles, paths of dynamic obstacles by gray lines and path found using our approach by white lines.}
  \label{fig:alg}
\end{figure*}

In our target problem of planning in a dynamic environment,
the low-dimensional state-space $S^{ld}$ consists of only spatial state variables, e.g., xy-coordinates, and the high-dimensional state-space $S^{hd}$ consists of states with spatio-temporal variables including a time dimension. Theoretically, the time dimension is unbounded and thus, the high-dimensional graph $G^{hd}$ is an infinite graph. For practical purposes, we bound the time dimension by a upper bound $T$, which would slightly modify the goal of planning problem into finding a least-cost path that can reach the goal from start within time $T$.
For any high dimensional state $X^{hd} \in S^{hd}$, we will use the notation $t(X^{hd})$ to denote the value of time dimension associated with that state. 

The projection function $\lambda$ projects the high-dimensional state $X^{hd}$ to a low-dimensional state $X^{ld}$ with only the spatial variables. 
If we follow the original definition of $\lambda$ in Equation~\ref{eq:lambda} then, for a given low-dimensional state $X^{ld}$, the inverse projection function $\lambda^{-1}$ would map state $X^{ld}$ to the set of all $X^{hd}$ where the spatial configuration of $X^{hd}$ is the same as $X^{ld}$ and $0 \leq t(X^{hd}) \leq T$. 
Thus, for each low-dimensional state, there are $T$ corresponding high-dimensional states, which is quite a large number as $T$ is usually a high value. 

Here, we introduce a pruning technique based on an observation that not all of high-dimensional states are reachable from the start state. For example, consider a low-dimensional state $X \in S^{ld}$ which is mapped to $T$ high-dimensional states. If the time-optimal path from start to this state, ignoring dynamic obstacles, reaches at time $t_f$ then all the states $X^{hd}$ with $\lambda(X^{hd}) = X$ and $t(X^{hd}) < t_f$ are essentially unreachable and hence, can be pruned away from the search space. 

Taking advantage of this fact, we decrease the size of search space by performing a low-dimensional time-optimal Dijkstra search in $G^{ld}$, which ignores dynamic obstacles, initially from the start state to all low-dimensional states and keep track of the time at which we reach each state. 
We store this time as a dependent variable $t_{dep}$ of the low-dimensional state and ignore all the corresponding high-dimensional states whose time value is less than $t_{dep}$ in the inverse projection mapping. 
Note that this dependent time variable need not be the exact optimal time obtained from the Dijkstra search, it just needs to be a lower bound on the optimal time. Pruning the search space in this way is necessary as it speeds up the planning by a considerable amount while still maintaining the completeness property of the planning algorithm.

Thus, we define the inverse projection function, $\lambda^{-1}$ as:
$$\lambda^{-1}(X^{ld}) = \{X^{hd} ~|~ \lambda(X^{hd}) = X^{ld}, t_{dep} \leq t(X^{hd}) \leq T\}$$
where $t_{dep}$ is the dependent time variable associated with the low-dimensional state $X^{ld}$.

The low-dimensional transition set is $T^{ld}=\{(X_i,X_j)|X_i,X_j \in S^{ld}\}$ where it is feasible for the robot to move from the spatial configuration of $X_i$ to $X_j$ according to its motion model. 
The transition set $T^{hd} = \{(X_i, X_j)|X_i, X_j \in S^{hd}\}$ where, $t(X_j) \geq t(X_i)$ and it is feasible for the robot to move from spatial configuration of $X_i$ to $X_j$ in time $(t(X_j) - t(X_i))$ according to its motion model. 
Note that we can check for collisions with any dynamic obstacle only in the high-dimensional transitions as we have the time information.

\subsection{Algorithm}
The planning algorithm follows that of~\cite{gochev2011path}. Here, we sketch the general algorithm and describe how it has been applied to our target problem of handling dynamic obstacles. 

\subsubsection{Adaptive Dimensionality Graph Construction}

The algorithm iteratively constructs $S^{ad}$, starting with $S^{ld}$ and introducing high-dimensional regions in subsequent iterations. Once a high-dimensional region is introduced we replace all the low-dimensional states that fall inside it with their high-dimensional counterparts as given by $\lambda^{-1}$ to get the re-constructed $S^{ad}$ for the next iteration. The transition set $T^{ad}$ is also iteratively constructed, starting with $T^{ld}$ and re-constructed as follows in subsequent iterations. 
For any state $X_i \in S^{ad}$:
\begin{itemize}
\item If $X_i$ is high-dimensional then, for all high-dimensional transitions $(X_i,X_j^{hd}) \in T^{hd}$, if $X_j^{hd} \in S^{ad}$ then $(X_i,X_j^{hd}) \in T^{ad}$. Otherwise, $(X_i, \lambda(X_j^{hd})) \in T^{ad}$.
\item If $X_i$ is low-dimensional then, for all low-dimensional transitions $(X_i, X_j^{ld}) \in T^{ld}$, if $X_j^{ld} \in S^{ad}$ then $(X_i, X_j^{ld}) \in T^{ad}$, and for all high-dimensional transitions $(X, X_j^{hd}) \in T^{hd}$ where $X \in \lambda^{-1}(X_i)$, if $X_j^{hd} \in S^{ad}$ then $(X_i, X_j^{hd}) \in T^{ad}$.
\end{itemize}

\subsubsection{Main Loop}
We start with $G^{ad}$ same as $G^{ld}$ and a high-dimensional region added at the start, which is necessary as the start state $X_S$ is high-dimensional with $t(X_S) = 0$. Note that the goal state $X_G$ is not high-dimensional as we don't know the value of the time dimension for the goal state. 

{\bf AD planning phase.} At the start of each iteration, the current graph $G^{ad}$ is searched for a path $\pi_{G^{ad}}^*$ from the start to the goal, using a suboptimal graph search algorithm like weighted A* with a suboptimality bound $\epsilon_{plan}$. During the search for this path, we consider the dynamic obstacles only in high-dimensional regions of $S^{ad}$ and not in the low-dimensional regions. Hence, the path found could potentially be in collision with a dynamic obstacle in the low-dimensional regions. If no path $\pi_{G^{ad}}^*$ is found, we return that there exists no feasible path that can reach the goal from start within time $T$, and the algorithm terminates.

{\bf Tracking phase.} If a path is found, then in the tracking phase, a high-dimensional tunnel $\tau$ is constructed around the path $\pi_{G^{ad}}^*$ and searched for the least-cost path $\pi_\tau^*(X_S, X_G^{hd})$ where $X_G^{hd} \in \lambda^{-1}(X_G)$. The tunnel is constructed by projecting all the states within to their high-dimensional counterparts. Notice that since the tunnel is entirely high-dimensional and is a subgraph of $G^{hd}$, we consider dynamic obstacles in the entire tunnel and hence, the path found is guaranteed to be feasible and collision-free. If a path $\pi_\tau^*$ is found and its cost is less than $\epsilon_{track}*c(\pi_{G^{ad}}^*)$, then it is returned as the solution by the algorithm and the algorithm terminates. If no path is found, then we identify the farthest location in the tunnel until which the planner has progressed (i.e. the path with most progress), introduce a high-dimensional region there and move onto next iteration. If a path is found and its cost is greater than $\epsilon_{track}*c(\pi_{G^{ad}}^*)$, then we identify the location where the largest cost discrepancy (cost difference) between the path $\pi_\tau^*$ and $\pi_{G^{ad}}^*$ is observed and a high-dimensional region is introduced there. In both cases, if we identify a location which is already high-dimensional, then the size of the high-dimensional region at that location is increased.

{\bf Graph updating phase.} The algorithm re-constructs $G^{ad}$ based on the new high-dimensional regions introduced and moves onto the next iteration of planning and tracking, and keeps repeating until it finds a feasible, collision-free path or returns that there is no such path. An example run of the algorithm is shown in Figure \ref{fig:alg}. The algorithm is presented in Algorithm \ref{alg:adplanning}.

\begin{algorithm}[h]
\begin{algorithmic}[1]
\State $G^{ad} = G^{ld}$
\State AddFullDimRegion($G^{ad}$, $\lambda(X_S)$)
\Loop
\State Search $G^{ad}$ for the path $\pi_{G^{ad}}^*(X_S, X_G)$
\If{no $\pi_{G^{ad}}^*(X_S, X_G)$ is found}
\State \Return no path from $X_S$ to $X_G$ within time $T$ exists
\EndIf
\State Construct tunnel $\tau$ around $\pi_{G^{ad}}^*(X_S, X_G)$
\State Search $\tau$ for the least-cost path $\pi_{\tau}^*(X_S, X_G^{hd})$ where $X_G^{hd} \in \lambda^{-1}(X_G)$
\If{no $\pi_{\tau}^*(X_S, X_G^{hd})$ is found}
\State Let $\pi(X_S, X_{end})$ be the path with most progress
\If{$X_{end}$ is high-dimensional}
\State GrowFullDimRegion($G^{ad}$, $\lambda(X_{end})$)
\Else
\State AddFullDimRegion($G^{ad}$, $X_{end}$)
\EndIf
\ElsIf{$c(\pi_{\tau}^*(X_S, X_G^{hd}))\textgreater\epsilon_{track}*c(\pi_{G^{ad}}^*(X_S, X_G))$}
\State Identify state $X_r$ with largest cost discrepancy
\If{$X_r$ is already high-dimensional}
\State GrowFullDimRegion($G^{ad}$, $\lambda(X_r)$)
\Else
\State AddFullDimRegion($G^{ad}$, $X_r$)
\EndIf
\Else
\State \Return $\pi_{\tau}^*(X_S, X_G^{hd})$
\EndIf
\EndLoop
\end{algorithmic}
\caption{Planning with AD in dynamic environments}
\label{alg:adplanning}
\end{algorithm}

\section{Theoretical Properties}
\label{theory}
The given algorithm is complete with respect to the underlying graph $G^{hd}$ and provides a sub-optimality bound on the cost of the returned path. \\ \\
\textbf{Theorem 5.1} \textit{If a path $\pi_\tau^*(X_S,X_G^{hd})$ is found in the tracking phase, it is guaranteed to be collision-free with respect to all obstacles} \\
\textbf{Proof} The tunnel $\tau$ constructed around $\pi_{G^{ad}}^*$ is entirely high-dimensional and is a subgraph of $G^{hd}$ therefore, the search space considers the transition set $T^{hd}$. In $T^{hd}$, transitions that are in collision with dynamic obstacles are assigned infinite cost, essentially making them invalid.

Hence, the path found in the tunnel $\pi_\tau^*$ is guaranteed to be collision-free with respect to all obstacles \QEDB \\ \\
\textbf{Theorem 5.2} \textit{If no path $\pi_{G^{ad}}^*(X_S,X_G)$ is found in the planning phase, then no collision-free, feasible path exists from start to goal in $G^{hd}$ that can reach the goal from the start within time $T$} \\
\textbf{Proof} If no path is found during the planning phase of the first iteration, no feasible path exists in the absence of dynamic obstacles (Note that we only consider dynamic obstacles in the high-dimensional regions during the planning phase). Therefore, there is no collision-free path. If no path is found during the planning phase of any subsequent iteration, then the algorithm was not able to progress in a high-dimensional region. It cannot be a low-dimensional region because in such a case, it would have terminated in a previous iteration. If the algorithm is not able to progress in a high-dimensional region, then all the transitions into or inside the region are blocked by dynamic obstacles. Because we allow transitions to all $X^{hd}$ inside the region where $t_{dep} \leq t(X^{hd}) \leq T$ and we already know that the planner cannot reach $X^{hd}$ at a time earlier than $t_{dep}$, that guarantees that there exists no path in $G^{hd}$ that can reach the goal from start within time $T$.\QEDB\\ \\
\textbf{Theorem 5.3} \textit{The algorithm always terminates after a finite number of iterations} \\
\textbf{Proof} If no path is found at the end of a iteration, we introduce either a new high-dimensional region or increase the size of an existing one. As the time dimension is bounded above by $T$, we have a finite state-space. Hence, in the worst scenario, after a finite number of iterations $G^{ad}$ will be the same as $G^{hd}$ and the algorithm will either terminate with a feasible path or return that there is no feasible, collision-free path. \QEDB \\\\
\iftrue
\textbf{Theorem 5.4} \textit{If a path $\pi(X_S, X_G^{hd})$ is found at the time of termination, its cost is no more than $\epsilon_{plan}*\epsilon_{track}*c(\pi_{G^{hd}}^*(X_S,X_G^{hd}))$ where $\pi_{G^{hd}}^*(X_S,X_G^{hd})$ is the optimal least-cost path in $G^{hd}$.} \\
\textbf{Proof} We obtain this bound using equation \ref{eq:cost} and the proof is similar to that of the AD approach. For this proof, we refer the reader to \cite{gochev2011path}. \QEDB 
\fi

\begin{table*}[t]
\tiny
\centering

\begin{tabular}{|c|c|c|c c|c c|c c|c c|}
\hline
Algorithm & Number of Success & Epsilon & \multicolumn{2}{|c|}{time (secs)} & \multicolumn{2}{|c|}{\# 4D expands} & \multicolumn{2}{|c|}{\# 2D expands} & \multicolumn{2}{|c|}{path cost} \\ 
 & & & mean & std dev & mean & std dev & mean & std dev & mean  & std dev \\ \hline\hline
Adaptive & 49 & 1.1 & 6.4 & 0.7 & 3160 & 1105 & 10810 & 9361 & 39442 & 4438 \\
\hline
4D & 7 & 1.1 & 99.3 & 67.7 & 127393 & 87024 & 0 & 0 & 37142 & 5766 \\
\hline\hline
Adaptive & 50 & 1.5 & 20.9 & 48.5 & 16688 & 42804 & 33276 & 88880 & 55342 & 14668 \\
\hline
4D & 40 & 1.5 & 67.1 & 75.8 & 85324 & 96306 & 0 & 0 & 49150 & 11568 \\
\hline\hline
Adaptive & 50 & 2.0 & 18.4 & 39.2 & 16029 & 42418 & 23193 & 62865 & 60050 & 17148 \\
\hline
4D & 44 & 2.0 & 36.5 & 61.5 & 45172 & 76970 & 0 & 0 & 54090 & 15339 \\
\hline
\end{tabular}

\caption{Results on 50 indoor environments with 10 dynamic obstacles.}
\label{tab:indoor-10}
\end{table*}

\begin{table*}[t]
\tiny
\centering

\begin{tabular}{|c|c|c|c c|c c|c c|c c|}
\hline
Algorithm & Number of Success & Epsilon & \multicolumn{2}{|c|}{time (secs)} & \multicolumn{2}{|c|}{\# 4D expands} & \multicolumn{2}{|c|}{\# 2D expands} & \multicolumn{2}{|c|}{path cost} \\ 
 & & & mean & std dev & mean & std dev & mean & std dev & mean  & std dev \\ \hline\hline
Adaptive & 41 & 1.1 & 6.7 & 0.8 & 3705 & 1379 & 12524 & 9901 & 40740 & 2200 \\
\hline
4D & 5 & 1.1 & 91.0 & 71.2 & 111165 & 87228 & 0 & 0 & 38320 & 6522 \\
\hline\hline
Adaptive & 40 & 1.5 & 11.7 & 14.0 & 7318 & 9285 & 21454 & 38559 & 54690 & 16811 \\
\hline
4D & 21 & 1.5 & 70.3 & 86.7 & 88576 & 109859 & 0 & 0 & 47566 & 9916 \\
\hline\hline
Adaptive & 46 & 2.0 & 18.5 & 26.6 & 16000 & 31148 & 13672 & 21383 & 57760 & 19450 \\
\hline
4D & 23 & 2.0 & 35.8 & 69.8 & 43546 & 86677 & 0 & 0 & 50039 & 12256 \\
\hline
\end{tabular}

\caption{Results on 50 indoor environments with 30 dynamic obstacles.}
\label{tab:indoor-30}
\end{table*}
\section{Experiments}
\label{exp}
For an experimental evaluation of the presented approach we use the domain of robotic path planning in dynamic environments for 3D-$(x,y,\theta)$ non-holonomic robot. To successfully avoid dynamic obstacles in the environment, we will need to add the time dimension to the state-space while planning. Hence, the full-dimensional planning has a 4D $(x,y,\theta,t)$ state-space. Our implementation of the algorithm kept track of the high-dimensional regions in the environment as spheres: 2D $(x,y)$ circles, in the 4D planning case, as this allowed to quickly check if a state falls inside a region or not, and also quickly add new regions or grow existing regions. 

We modeled our environment as a planar world and the robot as a polygonal object with a unicycle motion model, which doesn't allow waiting in place actions. Our adaptive-dimensionality space consists of a 2D $(x,y)$ low-dimensional state-space and a 4D $(x,y,\theta,t)$ high-dimensional state-space, where $\theta$  is the heading of the robot. Thus the projection function is:
$$ \lambda(x,y,\theta,t) = (x,y)$$
We used a set of 16-discretized values for the heading angle and a maximum value of $T = 1000$ seconds at a resolution of 0.1 seconds for the time dimension. The set of motion primitives used for the 4D states consists of pre-computed kinematically feasible motion sequences as used in a lattice-type planner \cite{likhachev2009planning}. The motion primitives used for the 2D states were the eight neighboring states according to the eight-connected 2D grid. Note that the motion primitives for 2D states do not produce feasible paths that can be executed by the robot. The objective of the planner is to find the minimal time path from the start state to goal state. Hence, cost of each edge in the graph is the time taken to execute the respective action multiplied by a constant factor.

\begin{figure}[ht]
  \centering
  \begin{subfigure}[t]{0.45\linewidth}
    \centering
    \includegraphics[width=\linewidth]{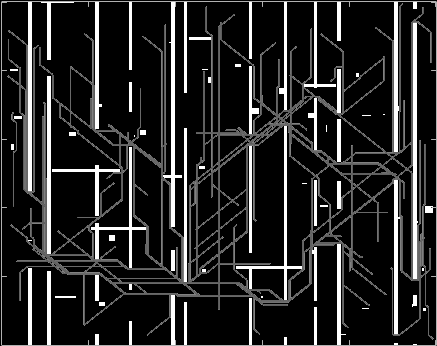}
    \caption{Maze-like environment}
  \end{subfigure}
  ~
  \begin{subfigure}[t]{0.45\linewidth}
    \centering
    \includegraphics[width=\linewidth]{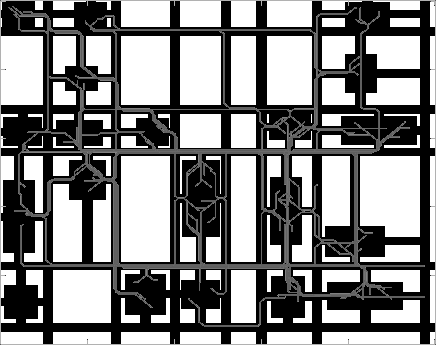}
    \caption{Indoor environment}
  \end{subfigure}
  \caption{Example maps, with paths (gray) of dynamic obstacles shown, used in our experiments. Static obstacles are shown in white and free space in black.}
  \label{fig:env}
\end{figure}
 
We compared our algorithm to the baseline 4D HCA* planner on several different environment sizes. In small environments with a few hundred cells along each spatial dimension, the baseline planner comes up with the plan quickly, so there is no advantage from our approach. In very large environments with more than 4000 cells along each spatial dimension, the baseline planner runs out of memory to find a solution, while our approach, since it deals with a low-dimensional state-space, was still able to plan successfully. To effectively compare the two approaches at the same level, we chose a moderate environment size of 2500 cells along each spatial dimension and generated 50 maze-like random environments like the one shown on the left in Figure \ref{fig:env}. We also generated 50 random indoor environments of the same size like the one shown on the right in Figure \ref{fig:env}. These indoor environments are composed of a series of narrow hallways and rooms on a grid placed randomly, while the maze-like environments are composed of a series of walls with small gaps in them to allow the robot to pass through.

In each of these environments, we randomly generate 30 dynamic obstacles. Each dynamic obstacle could come in a large or small size, randomly chosen, and started at a random configuration in the environment. To generate the trajectory for a dynamic obstacle, random goals were chosen and 2D A* is used to generate the paths between the goal points. We chose the start and goal configuration for each dynamic obstacle so that the resulting path is long enough, ensuring that they cover a significant area of the environment. In the indoor environments, the large dynamic obstacles fill the entire width of the hallways, so there is no way to pass them while the narrow dynamic obstacles fill half the width of the hallway, so they can be passed. In the maze-like environments, the dynamic obstacles traverse through the small gaps that the robot tries to pass through, resulting in congestion at such gaps. For each set of environments, we execute two sets of runs - one with 10 dynamic obstacles in each environment and the other with an additional 20 dynamic obstacles (making it a total of 30 moving obstacles) in each environment.

The underlying search algorithm used in both the planning and tracking phase is weighted A* with the $\epsilon_{plan}$ sub-optimality bound. The tunnel width used for the tracking phase, was 10 cells, and the radii of the newly added spheres were 20 cells. For the heuristic used by the weighted A* planners, in our approach and the baseline HCA*, we use a 2D Dijkstra search from the goal state to all the $(x,y)$ cells in the environment ignoring the dynamic obstacles. During the computation of heuristic, static obstacles are inflated by the inscribed circle radius of the robot to preclude paths through areas that are too narrow for the robot to physically traverse. 

For each environment, we try three values of $\epsilon$: 1.1, 1.5, and 2 with the adaptive planner using the square root of $\epsilon$ for both $\epsilon_{plan}$ and $\epsilon_{track}$, thus giving an overall sub-optimality bound of $\epsilon$ for the adaptive planner. We use the same set of $\epsilon$ values for the baseline planner and compare their performance. For both planners, we enforce a maximum planning time of 5 minutes for all $\epsilon$ values. 


\begin{table*}[t]
\tiny
\centering

\begin{tabular}{|c|c|c|c c|c c|c c|c c|}
\hline
Algorithm & Number of Success & Epsilon & \multicolumn{2}{|c|}{time (secs)} & \multicolumn{2}{|c|}{\# 4D expands} & \multicolumn{2}{|c|}{\# 2D expands} & \multicolumn{2}{|c|}{path cost} \\ 
 & & & mean & std dev & mean & std dev & mean & std dev & mean  & std dev \\ \hline\hline
Adaptive & 47 & 1.1 & 7.9 & 1.4 & 5105 & 2177 & 26665 & 9660 & 432425 & 43683 \\
\hline
4D & 4 & 1.1 & 161.5 & 59.8 & 195758 & 79955 & 0 & 0 & 416650 & 40272 \\
\hline\hline
Adaptive & 48 & 1.5 & 14.2 & 11.2 & 9622 & 8566 & 22588 & 21539 & 532652 & 91234 \\
\hline
4D & 45 & 1.5 & 41.3 & 43.8 & 51515 & 56488 & 0 & 0 & 562109 & 95470 \\
\hline\hline
Adaptive & 48 & 2.0 & 12.6 & 16.6 & 11771 & 13531 & 25630 & 34279 & 537873 & 89790 \\
\hline
4D & 48 & 2.0 & 18.6 & 35.8 & 22485 & 45048 & 0 & 0 & 622739 & 104136 \\
\hline
\end{tabular}

\caption{Results on 50 maze-like environments with 10 dynamic obstacles.}
\label{tab:maze-10}
\end{table*}

\begin{table*}[t]
\tiny
\centering

\begin{tabular}{|c|c|c|c c|c c|c c|c c|}
\hline
Algorithm & Number of Success & Epsilon & \multicolumn{2}{|c|}{time (secs)} & \multicolumn{2}{|c|}{\# 4D expands} & \multicolumn{2}{|c|}{\# 2D expands} & \multicolumn{2}{|c|}{path cost} \\ 
 & & & mean & std dev & mean & std dev & mean & std dev & mean  & std dev \\ \hline\hline
Adaptive & 46 & 1.1 & 18.2 & 16.1 & 12565 & 12843 & 50012 & 12470 & 441100 & 91923 \\
\hline
4D & 2 & 1.1 & 192.7 & 159.0 & 236515 & 196442 & 0 & 0 & 424250 & 81529 \\
\hline\hline
Adaptive & 48 & 1.5 & 25.4 & 33.8 & 16558 & 18703 & 29679 & 28739 & 524451 & 83024 \\
\hline
4D & 45 & 1.5 & 46.7 & 46.0 & 59116 & 59870 & 0 & 0 & 553686 & 89470 \\
\hline\hline
Adaptive & 48 & 2.0 & 22.9 & 36.2 & 18922 & 22002 & 44550 & 107434 & 538370 & 92375 \\
\hline
4D & 48 & 2.0 & 21.3 & 34.7 & 25986 & 43153 & 0 & 0 & 623997 & 103201 \\
\hline
\end{tabular}

\caption{Results on 50 maze-like environments with 30 dynamic obstacles.}
\label{tab:maze-30}
\end{table*}
\section{Results}
\label{results}
For both sets of environments, we compare the planning time, number of high-dimensional (4D) states expanded, number of low-dimensional (2D) states expanded and resulting path cost, between our approach and the baseline 4D HCA* approach. We also list out the number of cases among the set of 50 environments that our approach could come up with a solution within 5 minutes of planning time and the number of cases the baseline approach could. Note that statistics like mean planning time, number of HD expansions, number of LD expansions and path cost, are computed only on cases where both approaches could come up with a solution within 5 minutes.

Tables \ref{tab:indoor-10} and \ref{tab:indoor-30} present the results for 50 randomly generated indoor environments with 10 and 30 dynamic obstacles respectively. In these environments, the low-dimensional heuristic used is very often misleading as it cannot account for dynamic obstacles and leads the search into a blocked hallway. For $\epsilon=1.1$, the planning problem is hard and the baseline could solve only 5 environments with 30 dynamic obstacles (7 in the case of 10 dynamic obstacles). In comparison, our approach could solve 41 of the 50 environments with 30 dynamic obstacles (49 in the case of 10 dynamic obstacles) with a substantially smaller mean planning time. As the $\epsilon$ value increases, the planning problem becomes easier and performance of the baseline approach improves. Even in these easier cases, our approach has a comparable performance, if not better than that of baseline. Results across tables \ref{tab:indoor-10} and \ref{tab:indoor-30} show that our approach performs well even when the number of obstacles increases, whereas the performance of baseline degrades substantially.

The results for the 50 randomly generated maze-like environments with 10 and 30 dynamic obstacles are presented in tables \ref{tab:maze-10} and \ref{tab:maze-30}. These environments are characterized by tight turns and potential dynamic obstacle collisions at the gaps in the walls. In most cases, the robot would have to swerve around the obstacle to avoid collision. Hence, the resulting path doesn't deviate significantly from the one suggested by heuristic. From the results we can observe that when the planning problem is difficult (for example, when $\epsilon = 1.1$), our approach could solve a large number of cases (47 and 46 of 50) when compared to the baseline (4 and 2 of 50). But as the $\epsilon$ value increases and the planning problem becomes easier, performance of baseline quickly catches up with our approach and in one of the case, outperforms our approach as well ($\epsilon=2$ in the 30 dynamic obstacles case). But in most runs, our approach performs better than the baseline in mean planning time and the path cost. Results across tables \ref{tab:maze-10} and \ref{tab:maze-30} show that there is not as much decrease in the performance of baseline with increase in number of obstacles, when compared to the indoor environments.

\section{Discussion}
\label{sec:discussion}

Interestingly, in indoor environments our approach returns paths with higher costs (but still within the suboptimality bound) when compared to the baseline approach. This is due to the fast low-dimensional 2D planning used in our approach which when a hallway is blocked finds an alternative path through an adjacent hallway, even if it is against heuristic. In contrast, the baseline tries to go through the blocked hallway suggested by the heuristic by wasting time before and entering the hallway once the obstacle comes out. Hence, the path returned by baseline often has low cost.

Generally, in environments where dynamic obstacles do not block the path suggested by heuristic, the baseline approach is fast and often outperforms our approach. This is the case in maze-like environments where the robot has to just swerve around the obstacle to avoid it. Hence, we see a good performance of baseline in such cases. But in cases where the solution required a path significantly different from the one computed by heuristic, baseline performs poorly and our approach outperforms it. This is the case in indoor environments where the entire hallway is blocked by an obstacle and the planner has to find an alternative path which might be against the heuristic. In such environments, as the number of dynamic obstacles increases, the heuristic becomes less informative and performance of baseline degrades. Our approach circumvents this through its iterative nature and fast low-dimensional planning.


\section{Conclusion and Future Work}\label{sec:conclusion}

In this work we have presented a new approach to path planning in dynamic environments that doesn't make any assumptions on the robot's motion model, but still achieves significant speedups in planning time over heuristic-based A*.
Our algorithm builds on the previously-developed algorithm for path planning with adaptive dimensionality to explicitly decrease the dimensionality of the search space in an adaptive manner.
The algorithm plans in full dimensional state-space in regions of the environment where there is a potential dynamic obstacle collision and in low-dimensional state-space elsewhere, thereby obtaining quicker planning times.
 We have proven that our approach returns feasible, collision-free paths in dynamic environments with bounds on solution cost sub-optimality. As shown in our results, we outperform full-dimensional planning algorithms such as HCA* by a substantial margin in tasks like navigation of non-holonomic robot in dynamic environments. 

As a part of future work, we plan to verify performance of the algorithm on a real robot navigating in a realistic environment with dynamic obstacles. We are exploring the possibility of using an incremental planner to reuse the search information from previous iterations to speed up planning in subsequent iterations. Currently, the planning algorithm starts from scratch at the start of each iteration and does not reuse search tree from previous iterations. We are also interested in relaxing our assumption of complete knowledge regarding the trajectories of dynamic obstacles and handle uncertainty in the predicted trajectories within the algorithm.


\section*{Acknowledgments}
\label{sec:acknowledgements}
This work was conducted in part through collaborative participation in the Robotics Consortium sponsored by the U.S Army Research Laboratory under the Collaborative Technology Alliance Program, Cooperative Agreement W911NF-10-2-0016, and in part by ONR under MURI grant ``Reasoning in Reduced Information Spaces'' (no. N00014-09-1-1052). The views and conclusions contained in this document are those of the authors and should not be interpreted as representing the official policies, either expressed or implied, of the Army Research Laboratory of the U.S. Government. The U.S. Government is authorized to reproduce and distribute reprints for Government purposes notwithstanding any copyright notation herein. The authors would also like to thank Venkatraman Narayanan and Kalin Gochev for their valuable inputs, and the anonymous reviewers for their suggestions.

\bibliographystyle{aaai}
\bibliography{vemula-references}

\end{document}